\documentclass{IEEEoj}
\usepackage{cite}
\usepackage{amsmath,amssymb,amsfonts}
\usepackage{algorithmic}
\usepackage{graphicx,color}
\usepackage{textcomp}
\usepackage{array}
\usepackage[caption=false,font=normalsize,labelfont=sf,textfont=sf]{subfig}
\usepackage{url}
\usepackage{verbatim}
\usepackage{hyperref}
\usepackage{tikz}
\usetikzlibrary{positioning,fit,shapes.geometric,shadows}
\usepackage{pgfplots}
\pgfplotsset{compat=1.18}
\usepackage{siunitx}
\usepackage{booktabs}
\usepackage[table]{xcolor}
\usepackage{multirow}
\usepackage{makecell}
\usepackage{forest}
\usetikzlibrary{shadows, positioning, calc, shapes.multipart}

\definecolor{ieeecyan}{RGB}{0, 174, 239}

\usepackage{overpic}
\usepackage{pifont}
\usepackage{csquotes}
\usepackage{ragged2e}
\usepackage{float}

\def\BibTeX{{\rm B\kern-.05em{\sc i\kern-.025em b}\kern-.08em
    T\kern-.1667em\lower.7ex\hbox{E}\kern-.125emX}}
\AtBeginDocument{\definecolor{ojcolor}{cmyk}{0.93,0.59,0.15,0.02}}
\def\OJlogo{\vspace{-14pt}\includegraphics[height=28pt]{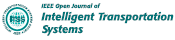}}

\begin{document}
\receiveddate{XX Month, XXXX}
\reviseddate{XX Month, XXXX}
\accepteddate{XX Month, XXXX}
\publisheddate{XX Month, XXXX}
\currentdate{XX Month, XXXX}
\doiinfo{OJITS.2022.1234567}

\title{A Comparative Evaluation of Large Vision-Language Models for 2D Object Detection under SOTIF Conditions}

\author{
    Ji Zhou$^{1,*}$,
    Yilin Ding$^{1,*}$,
    Yongqi Zhao$^{1}$,
    Jiachen Xu$^{2}$,
    Dong Bi$^{1, 3}$,
    Johannes Betz$^{4}$,
    and Arno Eichberger$^{1}$,~\IEEEmembership{Member,~IEEE}
}

\affil{Institute of Automotive Engineering, Graz University of Technology, 8010 Graz, Austria (e-mail: ji.zhou@student.tugraz.at; yilin.ding@student.tugraz.at; yongqi.zhao@tugraz.at; arno.eichberger@tugraz.at)}
\affil{Independent Researcher, 200000 Shanghai, China (e-mail: jcxu97@gmail.com)}
\affil{School of Intelligent Connected Vehicle, Hubei University of Automotive Technology, 442002 Shiyan, China (e-mail: dbi@huat.edu.cn)}
\affil{Professorship of Autonomous Vehicle Systems, Technical University of Munich, 85748 Garching, Germany (e-mail: johannes.betz@tum.de)}
\corresp{CORRESPONDING AUTHOR: Yongqi Zhao (e-mail: yongqi.zhao@tugraz.at).}
\authornote{$^{*}$Ji Zhou and Yilin Ding contributed equally to this work. The code of this work is available via the following link: \url{https://github.com/ftgTUGraz/LLM-SOTIF}.}

\markboth{A Comparative Evaluation of LVLMs for 2D Object Detection under SOTIF Conditions}{Zhou \textit{et al.}}

\begin{abstract}
Reliable environmental perception remains one of the main obstacles for safe operation of automated vehicles. Safety of the Intended Functionality (SOTIF) concerns safety risks from perception insufficiencies, particularly under adverse conditions where conventional detectors often falter. While Large Vision-Language Models (LVLMs) demonstrate promising semantic reasoning, their quantitative effectiveness for safety-critical 2D object detection is underexplored. This paper presents a systematic evaluation of ten representative LVLMs using the PeSOTIF dataset, a benchmark specifically curated for long-tail traffic scenarios and environmental degradations. Performance is quantitatively compared against two specialized detectors: the anchor-based YOLOv5 and the transformer-based RT-DETRv4. Experimental results reveal a critical trade-off: top-performing LVLMs (e.g., Gemini 3) surpass the YOLOv5 in recall by over 25\% and closely match RT-DETRv4 under natural visual degradation, while specialized detectors retain an advantage in geometric precision for handcrafted perturbations. These findings highlight the complementary strengths of semantic reasoning versus geometric regression, supporting the use of LVLMs as high-level safety validators in SOTIF-oriented automated driving systems.
\end{abstract}

\begin{IEEEkeywords}
Automated driving, Safety of the Intended Functionality, Vision-Language Models, Perception failure.
\end{IEEEkeywords}

\maketitle

\section{Introduction}

\begin{figure}[ht]
    \centering
    \includegraphics[width=.8\columnwidth]{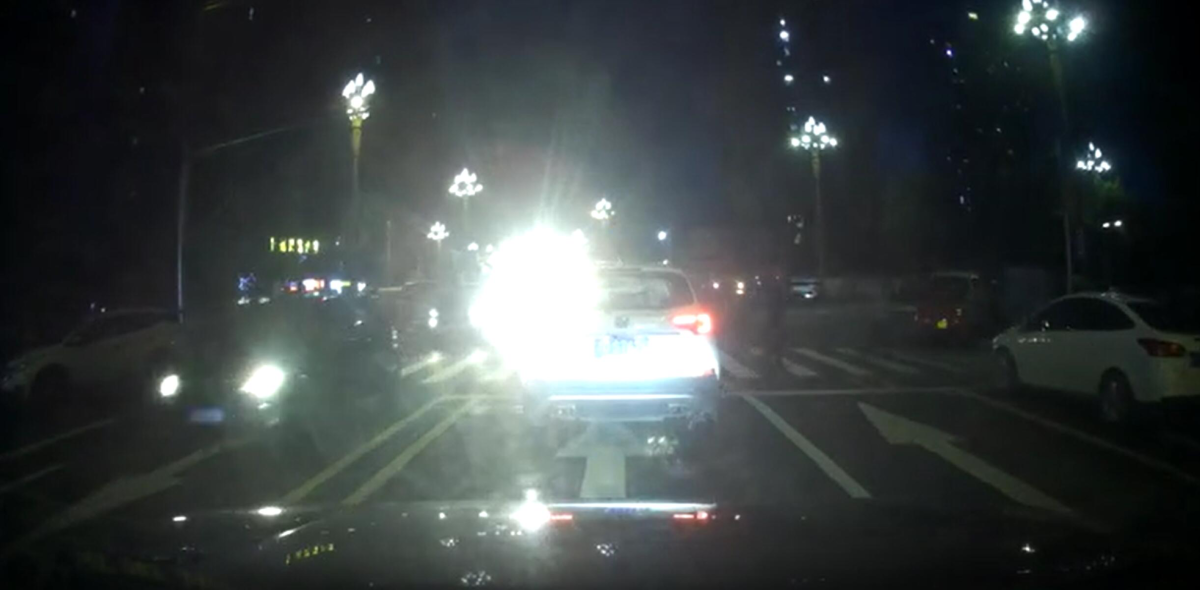}
    \caption{An example of potential perception functional insufficiency under severe glare from oncoming traffic, as provided in the PeSOTIF dataset~\cite{peng2023pesotif}.}
    \label{fig:PeSOTIF-Sample-Glare}
\end{figure}

\IEEEPARstart{A}{utomated} vehicles rely heavily on onboard perception systems to sense and interpret their surroundings, with vision-based sensing playing a crucial role in environment understanding~\cite{marti2019review}. Beyond failure caused by hardware or software malfunctions, safety risks may also arise when correctly functioning perception algorithms encounter conditions that exceed their operational capabilities, such as severe glare or low visibility. To address such non-malfunction-related safety risks, Safety of the Intended Functionality (SOTIF), formalized in ISO 21448, focuses on hazards caused by functional insufficiency rather than component faults~\cite{ISO21448_2022}. For example, as shown in~\autoref{fig:PeSOTIF-Sample-Glare}, severe glare from oncoming traffic can significantly degrade visibility. In such scenarios, an automated vehicle may fail to detect a pedestrian or another vehicle even though the perception systems operate as specified. Such failure modes highlight systematic perception limitations that are not attributable to component malfunctions.

Conventional object detection benchmarks in autonomous driving, such as KITTI~\cite{geiger2012we}, nuScenes~\cite{caesar2020nuscenes} and BDD100K~\cite{Yu_2020_CVPR}, are primarily designed around common traffic scenarios and standard detection conditions, and therefore often underrepresent perception failures in adverse environments. This discrepancy motivates SOTIF-oriented datasets that explicitly target non-malfunction-related perception limitations. To this end, PeSOTIF dataset~\cite{peng2023pesotif} was introduced to curate challenging scenarios, such as the one illustrated in~\autoref{fig:PeSOTIF-Sample-Glare}, characterized by adverse conditions and atypical objects, which are prone to inducing perception failures. By systematically capturing these corner cases, PeSOTIF serves as a dedicated testbed for identifying the operational boundaries of perception systems and evaluating their robustness against functional insufficiencies.

Motivated by the SOTIF-oriented perception failures discussed above, existing algorithmic efforts for object detection can be broadly grouped into two paradigms. One line of research enhances conventional detectors by explicitly enlarging safety margins under perceptual uncertainty, thereby improving robustness within the paradigm of detector-centric architectures. For example, Peng et al.~\cite{peng2021uncertainty} incorporate probabilistic inference mechanisms into the YOLO architecture to quantify uncertainty and identify perception risk in complex SOTIF scenarios. Similarly, Wang et al.~\cite{wang2025ensuring} augment YOLO~\cite{Redmon_2016_CVPR} with enhanced geometric constraints, achieving improved robustness and higher detection accuracy on KITTI.

In parallel, the emergence of Large Vision-Language Models (LVLMs) offers a new perspective: leveraging their vast world knowledge and semantic reasoning capabilities to handle open-set concepts and interpret complex scenarios. Representative studies focus on high-level reasoning, interpretation, or decision-making tasks using models such as GPT-4 and Video-LLaVA~\cite{10588373}, or propose SOTIF-oriented frameworks for risk interpretation specifically under adverse conditions~\cite{huang2025drivesotif}. Other efforts employ LLMs as auxiliary tools, for instance by guiding diffusion-based data augmentation to improve few-shot object detection~\cite{jiang2025llm}. However, these studies predominantly operate at the semantic level. It remains unverified whether the high-level reasoning of LVLMs can be effectively translated into precise geometric localization, a prerequisite for safe motion planning. Consequently, it is unclear how their detection performance compares systematically against established detectors such as YOLO on dedicated benchmarks like PeSOTIF.

To bridge this gap, the presented work conducts a comprehensive benchmark of ten mainstream LVLMs for object detection under SOTIF conditions, employing established YOLOv5 and RT-DETRv4~\cite{lv2024detrs} detectors as comparative baselines. The main contributions are summarized as follows:
\begin{enumerate}
    \item A unified evaluation pipeline is developed to enable LVLMs to perform 2D object detection via visual grounding under SOTIF conditions, effectively bridging the gap between high-level semantic reasoning and precise geometric localization without task-specific fine-tuning.
    \item A systematic benchmark of ten representative large foundation models is conducted for SOTIF-oriented object detection, using PeSOTIF as a dedicated testbed and two specialized detectors, YOLOv5 and RT-DETRv4, as reference baselines.
    \item A quantitative analysis of detection performance and failure patterns is provided, revealing a critical trade-off where LVLMs exhibit superior robustness in semantic recall but lag behind conventional detectors in geometric precision.
\end{enumerate}

% \begin{itemize}
%     \item A pipeline to enhance the visual grounding capability of LLMs.
%     \item Quantify the LLM's capability in 2D object detection.
% \end{itemize}

By establishing this benchmark, it is aimed to provide a quantitative reference for the safe deployment of foundation models in automated driving. The insights derived from the performance trade-offs are expected to guide the design of future hybrid perception systems, which combine the semantic robustness of LVLMs with the geometric precision of specialized detectors to address the safety challenges of SOTIF.

\section{Methodology}

\begin{figure*}[ht]
    \centering
    \includegraphics[width=\textwidth]{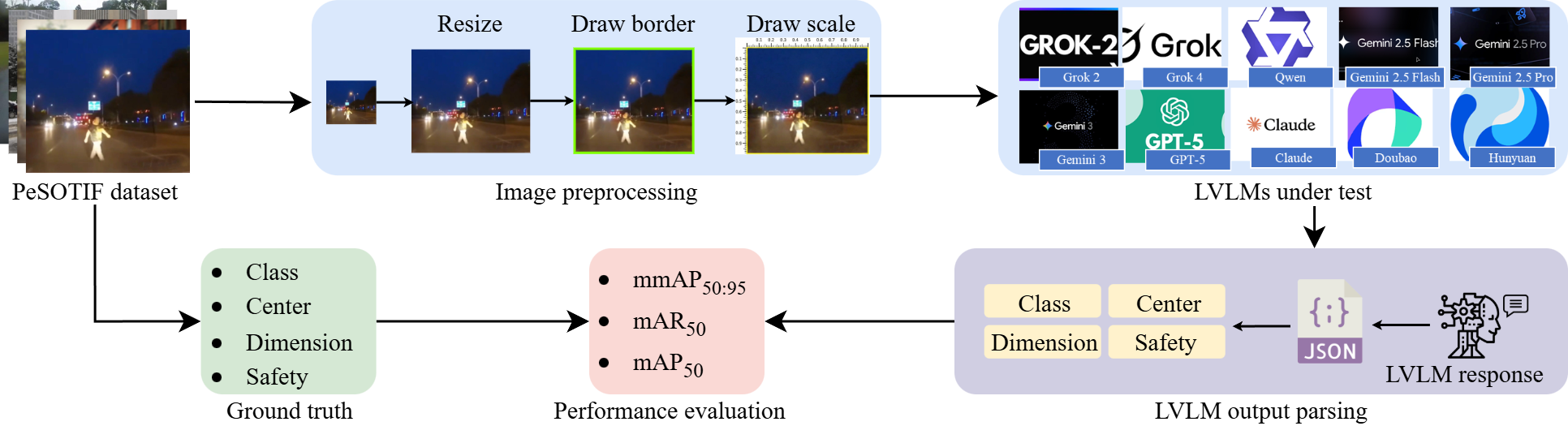}
    \caption{Overall workflow of the LVLM-based 2D object recognition in PeSOTIF dataset.}
    \label{fig:LLM_SOTIF}
\end{figure*}

The overall workflow of the presented work is depicted in~\autoref{fig:LLM_SOTIF}. The process initiates with image preprocessing, where raw samples from the PeSOTIF dataset~\cite{peng2023pesotif} are standardized through resizing, border annotation, and the superimposition of scale markers. Subsequently, the processed inputs are fed into ten LVLMs for object detection, with the generated responses parsed into a structured format consistent with the dataset annotations. In the final stage\footnote{The metrics are explained in \autoref{sec-performance-eva}.}, performance evaluation is conducted by quantitatively comparing the predicted bounding boxes against ground-truth labels to measure comprehensive detection performance.

\subsection{PeSOTIF Dataset}

This work employs the PeSOTIF dataset~\cite{peng2023pesotif} as the testbed. As a dedicated benchmark for SOTIF-oriented perception, PeSOTIF comprises 1126 frames capturing long-tail traffic scenarios. Distinct from conventional datasets, it prioritizes scenes characterized by perception-degrading conditions and atypical road anomalies, which serve as  stress tests for vision-based systems.

\begin{figure}[ht]
    \centering
    \includegraphics[width=\columnwidth]{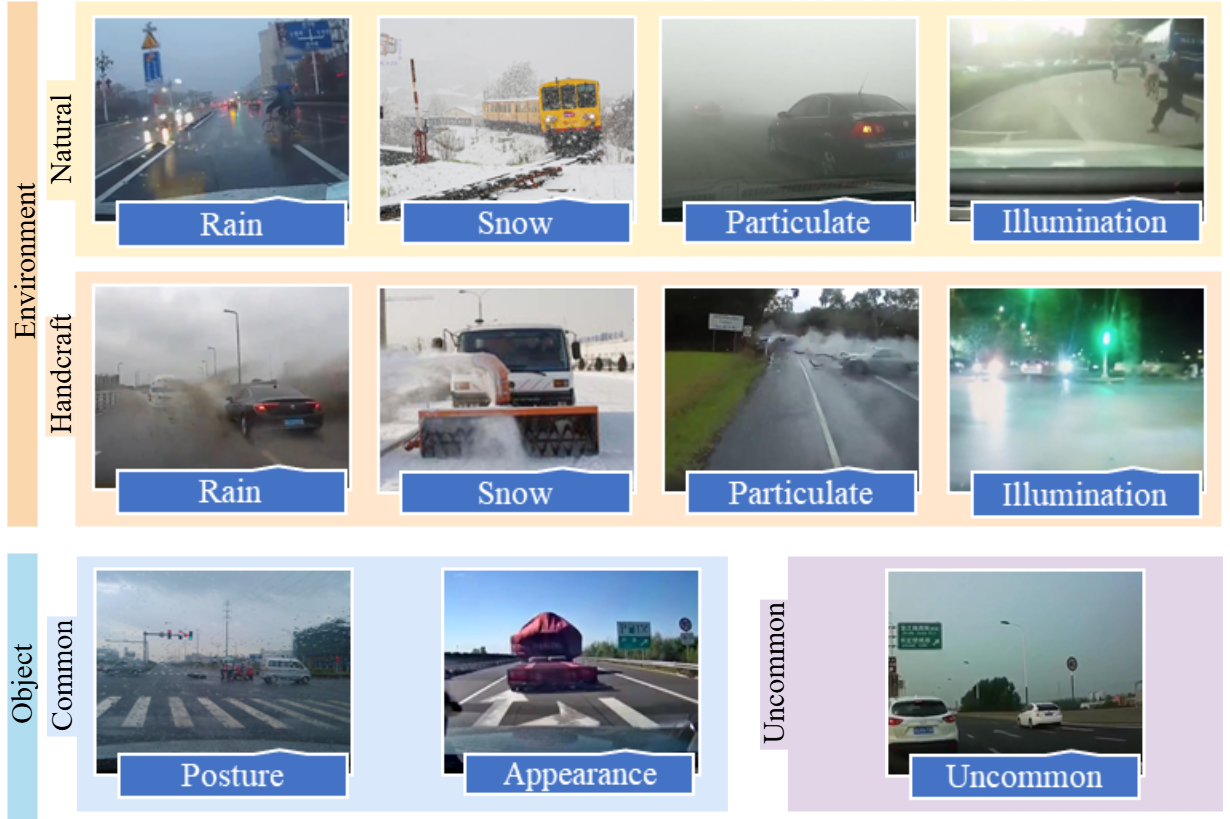}
    \caption{Overview of PeSOTIF dataset~\cite{peng2023pesotif}.}
    \label{fig:PeSOTIF_Dataset}
\end{figure}

As depicted in~\autoref{fig:PeSOTIF_Dataset}, PeSOTIF is structured into two subsets. The environment subset categorizes visual degradation into natural conditions (e.g., rain) and handcrafted perturbations (e.g., synthetic glare). Complementing this, the object subset distinguishes between common vehicles and atypical anomalies (e.g., overturned trucks).

\subsection{Image Preprocessing} 

\subsubsection{Image Resize}

As shown in the first stage of the preprocessing pipeline (\autoref{fig:LLM_SOTIF}), all images are resized using a standardized procedure to ensure a consistent input format across the dataset and to facilitate LVLM inference. Each raw image $I$, with original width $W_{\text{orig}}$ and height $H_{\text{orig}}$, was scaled to fit within bounds $W_{\text{max}}=800$ and $H_{\text{max}}=600$ pixels, thereby providing a uniform coordinate space for subsequent evaluation. The resizing operation preserved the original aspect ratio to avoid geometric distortion of traffic objects. A global scaling factor $S$ was computed as

\begin{equation}
S=\min\left(\frac{W_{\max}}{W_{\text{orig}}},\frac{H_{\max}}{H_{\text{orig}}}\right).
\end{equation}

The resized dimensions were then obtained by $W_{\text{new}}=S\cdot W_{\text{orig}}$ and $H_{\text{new}}=S\cdot H_{\text{orig}}$. To reduce aliasing\footnote{Aliasing describes distortions introduced during image resizing or downsampling, where fine spatial details are not correctly preserved.} during downsampling and maintain visual quality, the Lanczos resampling filter~\cite{turkowski1990filters} was applied.

\subsubsection{Draw Border} 
After resizing, a clear visual boundary was added to define the spatial extent of the input. Specifically, a continuous rectangular border with a linewidth of 2 pixels was drawn along the image perimeter, spanning from $(0, 0)$ to $(W-1, H-1)$. This boundary serves two purposes: it provides an explicit visual cue that delineates the Region of Interest (ROI) for the LVLM, and it helps separate scene content from the canvas edges, which may reduce coordinate estimation errors near image boundaries.

% \subsubsection{Add Ruler}
\subsubsection{Draw Scale}
Because LVLMs may be limited in precise spatial localization\footnote{This limitation is closely related to the concept of \emph{visual grounding}, which describes the alignment between language expressions and image regions in computer vision.}, an explicit visual coordinate reference was embedded into the image. A visual coordinate reference in the form of ruler ticks was overlaid along the top (horizontal) and left (vertical) margins of the image. The coordinate range was divided into ten equal intervals, with tick marks placed every 10\% of the image width ($W$) and height ($H$), respectively. Normalized numeric labels (e.g., 0.1, 0.2, ..., 0.9) were displayed next to each tick. This augmentation provides a direct spatial reference within the input. By acting as explicit visual landmarks, these ticks allow the model to \enquote{read} coordinates directly from the image rather than implicitly, thereby improving the precision of bounding-box regression in the normalized coordinate domain $[0,1]$.

\subsection{LVLMs Under Test}

\subsubsection{LVLMs Specification}
To ensure a comprehensive evaluation, ten state-of-the-art foundation models were selected from major developers, including Google, OpenAI, xAI, Anthropic, Alibaba, ByteDance, and Tencent. These models represent the forefront of multimodal capabilities as of late 2025. The selection covers a broad spectrum of parameter scales where publicly disclosed, ranging from lightweight models to trillion-parameter giants. The candidate models include: Grok 2, Grok 4, Gemini 2.5 Pro, Gemini 2.5 Flash, Gemini 3, GPT-5, Claude 4.5, Qwen 3-Max, Doubao and Hunyuan 2.0. Key specifications, including developer, release note, and parameter count, are summarized in~\autoref{tab:llm_specs}.

\begin{table}[ht]
\caption{Specifications of LVLMs under test}
\label{tab:llm_specs}
\centering
\small
\begin{tabular}{llll}
    \toprule[2pt]
    \textbf{Model} & \textbf{Developer} & \textbf{Release Date} & \textbf{Parameter} \\
    \midrule[1pt]
    Grok 2 & xAI & Aug. 2024 &270B \\
    Grok 4 & xAI & July 2025 & 1.7T \\
    Gemini 2.5 Pro & Google & June 2025 & 128B \\
    Gemini 2.5 Flash & Google & May 2025 & 5B \\
    Gemini 3 & Google & Nov. 2025 & - \\
    GPT-5 & OpenAI & Aug. 2025 & - \\
    Qwen 3-Max & Alibaba & Sep. 2025 & 1T \\
    Claude 4.5 & Anthropic & Nov. 2025 & - \\
    Doubao & ByteDance & May 2024 & - \\
    Hunyuan 2.0 & Tencent & Dec. 2025 & 32B \\
    \bottomrule[2pt]
\end{tabular}
\end{table}

\subsubsection{Prompt Engineering} As depicted in~\autoref{fig:prompt_LLM}, the prompt is designed with a structured chain-of-thought~\cite{NEURIPS2022_9d560961} approach, comprising four strategic components to maximize inference reliability: \textit{(a) Role Definition} establishes an expert persona for image-based perception, aiming to activate the model's domain-specific latent knowledge;~\textit{(b) Task Specification} explicitly details the 11-class taxonomy and instructs the model to use the visual rulers grounding, thereby aligning semantic understanding with the normalized coordinate space $[0, 1]$;~\textit{(c) Output Format} enforces a strict JSON schema for class IDs and bounding boxes, facilitating automated parsing;~\textit{(d) Constraints} restrict the response to the specified fields to ensure consistent, machine-readable outputs.

\begin{figure}[ht]
\centering
\footnotesize
% Ask the model to adopt a persona
\begin{minipage}[c]{0.05\columnwidth}
\centering
\subfloat{(a)}
\label{fig:personaBoxSub}
\end{minipage}%
\begin{minipage}[c]{0.93\columnwidth}
\begin{tikzpicture}
\node[draw, minimum width=0.93\columnwidth, align=justify, text width=0.91\columnwidth, fill=blue!15, draw=blue!60, line width=0.5pt, drop shadow={shadow xshift=0.5mm, shadow yshift=-0.5mm, opacity=0.3}] (personaBox) {
As an image recognition expert, your task is to analyze images from dashcam footage and provide output in JSON format with the following keys only: class, x\_center, y\_center, width, length and safety critical.
};
\end{tikzpicture}
\end{minipage}

\vspace{3pt}

% Descriptive text of scenario
\begin{minipage}[c]{0.05\columnwidth}
\centering
\subfloat{(b)}
\label{fig:sceDespBoxSub}
\end{minipage}%
\begin{minipage}[c]{0.93\columnwidth}
\begin{tikzpicture}
\node[draw, minimum width=0.93\columnwidth, align=justify, text width=0.91\columnwidth, fill=blue!15, draw=blue!60, line width=0.5pt, drop shadow={shadow xshift=0.5mm, shadow yshift=-0.5mm, opacity=0.3}](sceDespBox){
- class represents the class of an object which divided like this: 0 car, 1 bus, 2 truck, 3 train, 4 bike, 5 motor, 6 person, 7 rider, 8 traffic sign, 9 traffic light, and 10 traffic cone\\
- the coordinate of object in the given picture should be normalized to 0-1, with the reference point [0,0] at the top left corner and the reference point [1,1] at the bottom right corner of the picture\\
- x center and y center should represent the coordinates of the center of the detected object within the image\\
- width and length represent a bounding box that frames exactly the outline of one object\\
- safety critical represents whether the object is safety critical for the driver proceed\\
- to give you reference of coordinates, rulers with a marker every 1/10th of width and height are drawn on top and left of image and the whole picture is being framed
};
\end{tikzpicture}
\end{minipage}

\vspace{3pt}

% Include details in your query to get more relevant answers
% Specify the steps required to complete a task
% Use delimiters to clearly indicate distinct parts of the input
\begin{minipage}[c]{0.05\columnwidth}
\centering
\subfloat{(c)}
\label{fig:taskDetailSub}
\end{minipage}%
\begin{minipage}[c]{0.93\columnwidth}
\begin{tikzpicture}
\node[draw, minimum width=0.93\columnwidth, align=justify, text width=0.91\columnwidth, fill=blue!15, draw=blue!60, line width=0.5pt, drop shadow={shadow xshift=0.5mm, shadow yshift=-0.5mm, opacity=0.3}](taskDetailBox){
Each \{\} represents an object in the picture. Please adhere strictly to this output structure:\\
\texttt{[}
\{
  "class": int value,\\
  "x\_center": float value to two decimal places,\\
  "y\_center": float value to two decimal places,\\
  "width": float value to two decimal places,\\
  "length": float value to two decimal places,\\
  "safety\_critical": int value
\}
\texttt{]}
};
\end{tikzpicture}
\end{minipage}

\vspace{3pt}

\begin{minipage}[c]{0.05\columnwidth}
\centering
\subfloat{(d)}
\label{fig:exampleSub}
\end{minipage}%
\begin{minipage}[c]{0.93\columnwidth}
\begin{tikzpicture}
\node[draw, minimum width=0.93\columnwidth, align=justify, text width=0.91\columnwidth, fill=blue!15, draw=blue!60, line width=0.5pt, drop shadow={shadow xshift=0.5mm, shadow yshift=-0.5mm, opacity=0.3}] (exampleBox) {
Note: Do not include any additional data or keys outside of what has been specified.
};
\end{tikzpicture}
\end{minipage}

\caption{Structure of the LLM prompt for 2D object detection, consisting of (a) role definition, (b) task specification with class taxonomy and coordinate reference, (c) output format specification in JSON, and (d) constraints to enforce format consistency.}
\label{fig:prompt_LLM}
\end{figure}

\subsection{LVLM Output Parsing}

The unstructured, verbose textual responses of LVLMs were transformed into structured data for quantitative analysis. Using regular expressions, the system extracted the JSON-formatted detection result from the raw outputs while removing additional text (e.g., explanations). The extracted JSON was then parsed to obtain the required object attributes, including class labels, normalized bounding-box coordinates, and safety-critical indicators. This procedure ensures that only valid and standardized detection records are passed to the downstream Intersection-Over-Union (IoU) evaluation module.

\begin{figure*}[ht]
    \centering
    \includegraphics[width=\textwidth]{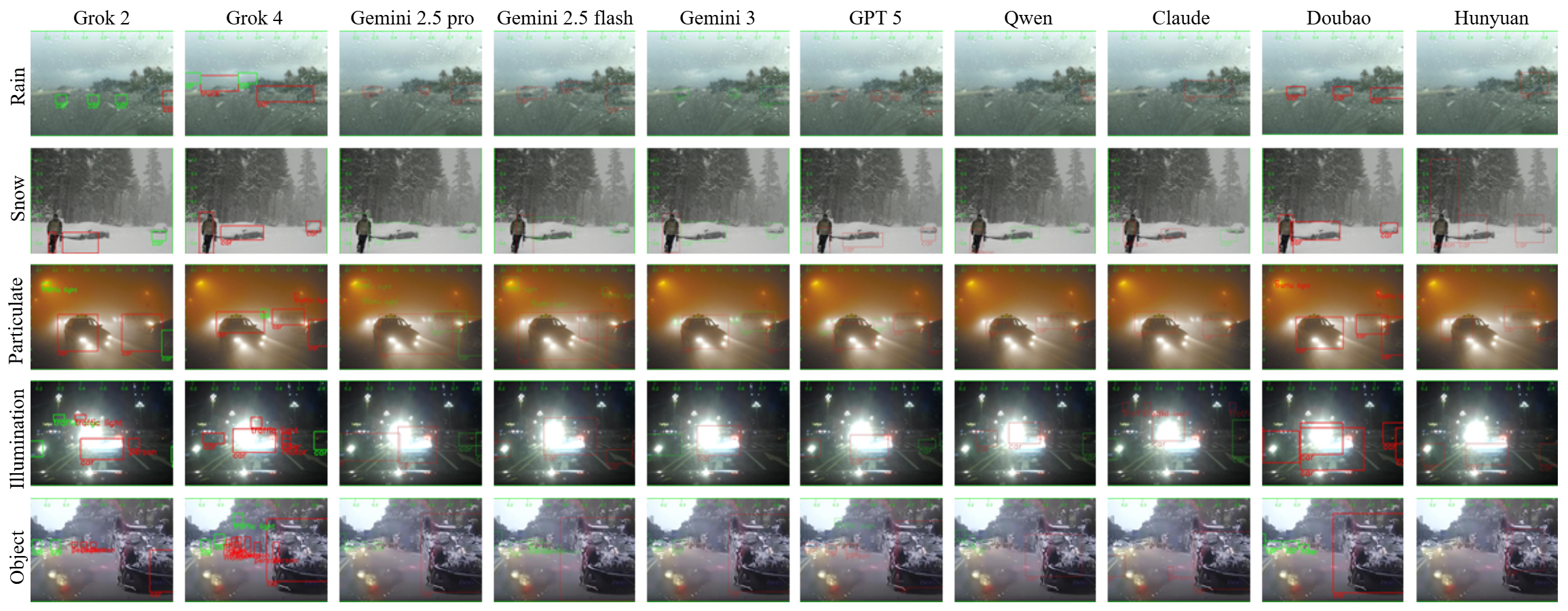}
    \caption{Qualitative visualization of ten LVLMs' performance in PeSOTIF dataset.}
    \label{fig:llm-performance-visual}
\end{figure*}

\subsection{Performance Evaluation}
\label{sec-performance-eva}

In this study, two specialized detectors are adopted as comparative baselines to span the two dominant paradigms of modern 2D object detection. First, YOLOv5 is selected as a representative of the single-stage, anchor-based convolutional detectors. While newer versions such as YOLOv11 exist, YOLOv5 is chosen for its recognized stability, extensive validation in safety-critical applications, and proven maturity as an industrial standard~\cite{jegham2025yoloevolutioncomprehensivebenchmark}. Specifically, the 60-epoch variant (YOLOv5$_{60e}$), reported in~\cite{peng2023pesotif} is adopted as the representative baseline. Second, RT-DETRv4~\cite{lv2024detrs} is introduced as a transformer-based, end-to-end detector that removes the hand-crafted non-maximum suppression step and achieves state-of-the-art speed-accuracy trade-offs on standard benchmarks. Including RT-DETRv4 alongside YOLOv5 enables the LVLMs to be compared against both the classical CNN lineage and the more recent DETR-style architectures that currently define the production frontier, thereby providing a more comprehensive reference for the benchmark. To assess LVLM performance in traffic-scene perception, standard object detection metrics based on IoU are adopted following the COCO benchmark protocols~\cite{10.1007/978-3-319-10602-1_48}. Detection precision is quantified using Mean Average Precision ($\text{mAP}_{50}$) at an IoU threshold of 0.50, which is the mean of Average Precision (AP) across all $N_c$ object classes:

\begin{equation}
    \text{mAP}_{50} = \frac{1}{N_c} \sum_{c=1}^{N_c} \text{AP}_c(0.50),
    \label{eq:map50}
\end{equation}

\noindent where $\text{AP}_c(0.50)$ denotes the average precision for class $c$. Similarly, the detection completeness is measured using Mean Average Recall ($\text{mAR}_{50}$), calculated as the mean of maximum recall values across classes:

\begin{equation}
    \text{mAR}_{50} = \frac{1}{N_c} \sum_{c=1}^{N_c} \text{AR}_c(0.50),
    \label{eq:mar50}
\end{equation}

\noindent where $\text{AR}_c(0.50)$ is the average recall for class $c$. Finally, to evaluate localization robustness across diverse overlap requirements, the component-averaged Mean Average Precision ($\text{mmAP}_{50:95}$) is computed by averaging the mAP over multiple IoU thresholds $\mathcal{T} = \{0.50, 0.55, \dots, 0.95\}$:

\begin{equation}
    \text{mmAP}_{50:95} = \frac{1}{|\mathcal{T}|} \sum_{t \in \mathcal{T}} \text{mAP}_t.
    \label{eq:mmap}
\end{equation}

\section{Result}

\autoref{fig:llm-performance-visual} visualizes the detection predictions across five representative SOTIF scenarios in the PeSOTIF dataset. As shown in the first three rows, general-purpose LVLMs demonstrate remarkable robustness under weather-induced degradation, effectively localizing objects even in rain, snow, and dense fog. Furthermore, these models exhibit strong semantic generalization in the last row, successfully grounding uncommon objects that challenge traditional detectors. These qualitative results suggest that LVLMs can translate high-level semantic understanding into reasonable spatial predictions, maintaining operation where traditional systems might fail.

\subsection{Overall Performance}

\begin{figure}[ht]
    \centering
    \begin{overpic}[width=.95\columnwidth]{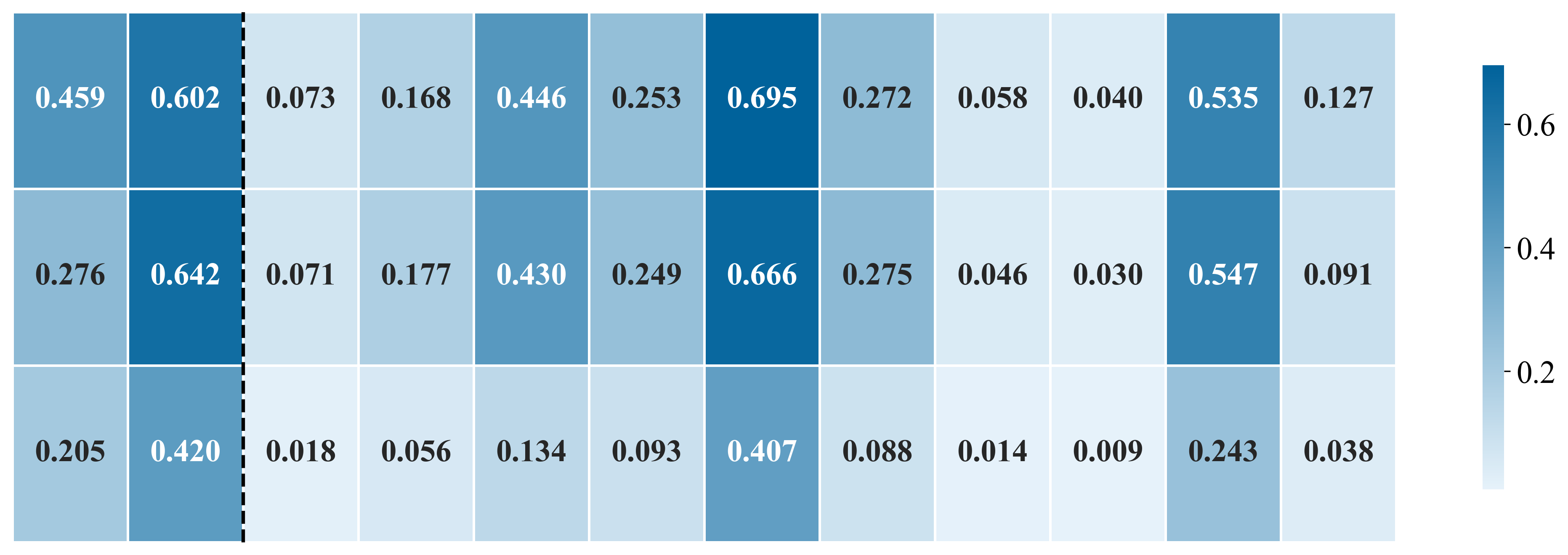}

        % --- X-Axis labels (12 columns: centers ~ 2, 9.5, 17, 24.5, 32, 39.5, 47, 54.5, 62, 69.5, 77, 84.5) ---
        \put(-1, -7){\rotatebox{45}{\scalebox{0.8}{\tiny YOLOv5$_{60e}$}}}
        \put(7, -7){\rotatebox{45}{\scalebox{0.8}{\tiny RT-DETRv4}}}
        \put(18, -3){\footnotesize \ding{172}}
        \put(25.5, -3){\footnotesize \ding{173}}
        \put(32.5, -3){\footnotesize \ding{174}}
        \put(40, -3){\footnotesize \ding{175}}
        \put(47.3, -3){\footnotesize \ding{176}}
        \put(54.5, -3){\footnotesize \ding{177}}
        \put(62, -3){\footnotesize \ding{178}}
        \put(69.5, -3){\footnotesize \ding{179}}
        \put(77, -3){\footnotesize \ding{180}}
        \put(84.5, -3){\footnotesize \ding{181}}

        % --- Y-Axis labels ---
        \put(-3, 5){\footnotesize \ding{182}}
        \put(-3, 16){\footnotesize \ding{183}}
        \put(-3, 28){\footnotesize \ding{184}}

    \end{overpic}

    \vspace{1em}

    \caption{Overall object-detection performance of LVLMs in PeSOTIF dataset, benchmarked against the YOLOv5$_{60e}$ and the RT-DETRv4~\cite{lv2024detrs} detectors. LVLMs:~\ding{172} Grok 2;~\ding{173} Grok 4;~\ding{174} Gemini 2.5 Pro;~\ding{175} Gemini 2.5 Flash;~\ding{176} Gemini 3;~\ding{177} GPT 5;~\ding{178} Qwen;~\ding{179} Claude;~\ding{180} Doubao;~\ding{181} Hunyuan. Metrics:~\ding{182} mmAP$_{50:95}$;~\ding{183} mAR$_{50}$;~\ding{184} mAP$_{50}$.}
    \label{fig:overall-performance}
\end{figure}

\autoref{fig:overall-performance} reports the overall object-detection performance of ten LVLMs compared to YOLOv5$_{60e}$ and RT-DETRv4~\cite{lv2024detrs}. RT-DETRv4 outperforms YOLOv5$_{60e}$ across all three metrics. Nevertheless, Gemini 3 (\ding{176}) achieves the highest mAP$_{50}$ score, followed by Doubao (\ding{180}) and Gemini 2.5 Pro (\ding{174}), with Gemini 3 surpassing both specialized detectors in mAP$_{50}$ and mAR$_{50}$, while falling marginally short of RT-DETRv4 in mmAP$_{50:95}$ (0.407 vs. 0.420). Moreover, the ranking is not fully explained by model parameter scale, as Gemini 2.5 Flash (\ding{175}, 5B) outperforms larger models, such as Grok 4 (1.7T) and Qwen 3-Max (1T), in terms of mAP$_{50}$, highlighting the role of architecture and training in spatially grounded prediction.

\subsection{Performance in Environment Subset}

\begin{figure}[ht]
    \centering
    \begin{overpic}[width=.95\columnwidth]{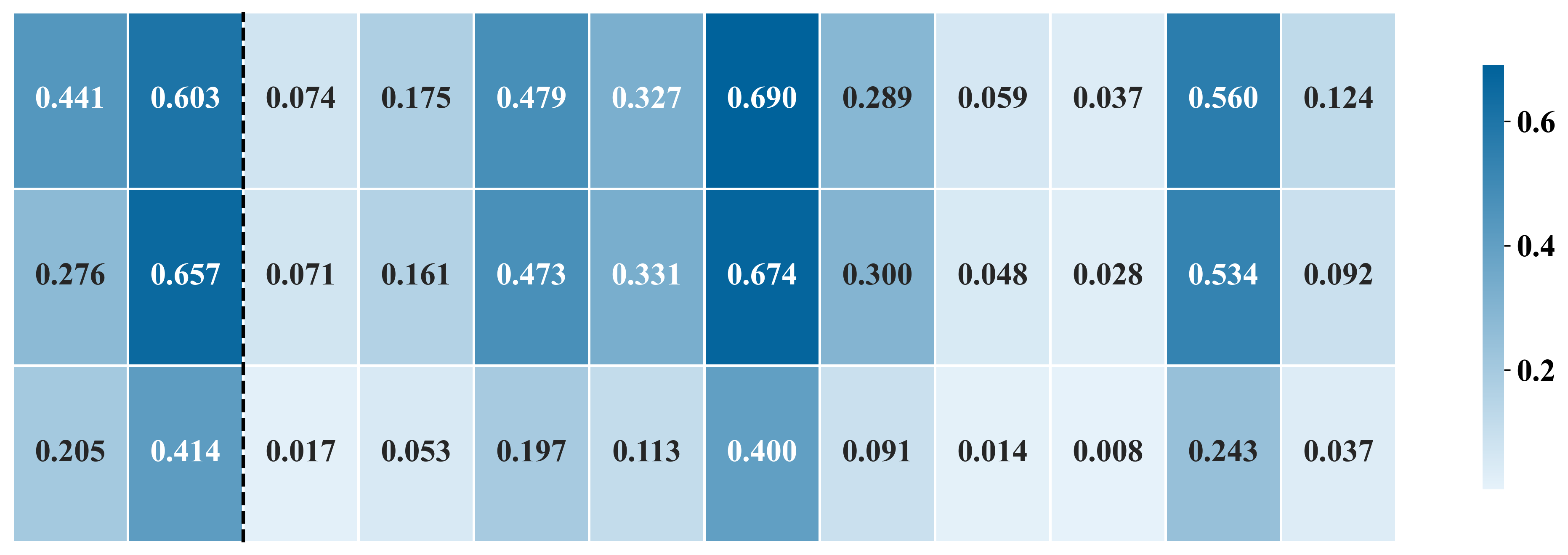}
        
        % --- X-Axis labels (12 columns: centers ~ 2, 9.5, 17, 24.5, 32, 39.5, 47, 54.5, 62, 69.5, 77, 84.5) ---
        \put(-1, -7){\rotatebox{45}{\scalebox{0.8}{\tiny YOLOv5$_{60e}$}}}
        \put(7, -7){\rotatebox{45}{\scalebox{0.8}{\tiny RT-DETRv4}}}
        \put(18, -3){\footnotesize \ding{172}}
        \put(25.5, -3){\footnotesize \ding{173}}
        \put(32.5, -3){\footnotesize \ding{174}}
        \put(40, -3){\footnotesize \ding{175}}
        \put(47.3, -3){\footnotesize \ding{176}}
        \put(54.5, -3){\footnotesize \ding{177}}
        \put(62, -3){\footnotesize \ding{178}}
        \put(69.5, -3){\footnotesize \ding{179}}
        \put(77, -3){\footnotesize \ding{180}}
        \put(84.5, -3){\footnotesize \ding{181}}

        % --- 纵轴 (Y-Axis) 设置 ---
        \put(-3, 5){\footnotesize \ding{182}}
        \put(-3, 16){\footnotesize \ding{183}}
        \put(-3, 28){\footnotesize \ding{184}}

    \end{overpic}
    
    \vspace{1em} 
    
    \caption{Performance of LVLMs in environment subset of PeSOTIF, benchmarked against the YOLOv5$_{60e}$ and the RT-DETRv4~\cite{lv2024detrs} detectors. LVLMs:~\ding{172} Grok 2;~\ding{173} Grok 4;~\ding{174} Gemini 2.5 Pro;~\ding{175} Gemini 2.5 Flash;~\ding{176} Gemini 3;~\ding{177} GPT 5;~\ding{178} Qwen;~\ding{179} Claude;~\ding{180} Doubao;~\ding{181} Hunyuan. Metrics:~\ding{182} mmAP$_{50:95}$;~\ding{183} mAR$_{50}$;~\ding{184} mAP$_{50}$.}
    \label{fig:env-subset-performance}
\end{figure}

\autoref{fig:env-subset-performance} summarizes detection performance on the environment subset of PeSOTIF, covering perception-degrading conditions such as rain, snow, particulates, and challenging illumination. Overall, several LVLMs remain competitive under these conditions. Gemini 3 (\ding{176}) achieves the best performance and surpasses both baselines in $mAP_{50}$ and $mAR_{50}$, while falling marginally short of RT-DETRv4 in $mmAP_{50:95}$. Doubao (\ding{180}) and Gemini 2.5 Pro (\ding{174}) also show strong results, outperforming YOLOv5$_{60e}$ but remaining below RT-DETRv4, indicating that LVLM-based approaches can better tolerate environmental interference in long-tail traffic scenes while only the strongest LVLM matches the transformer-based detector.

\subsubsection{Performance in Natural Subset}

\begin{figure}[ht]
    \centering
    \begin{overpic}[width=.95\columnwidth]{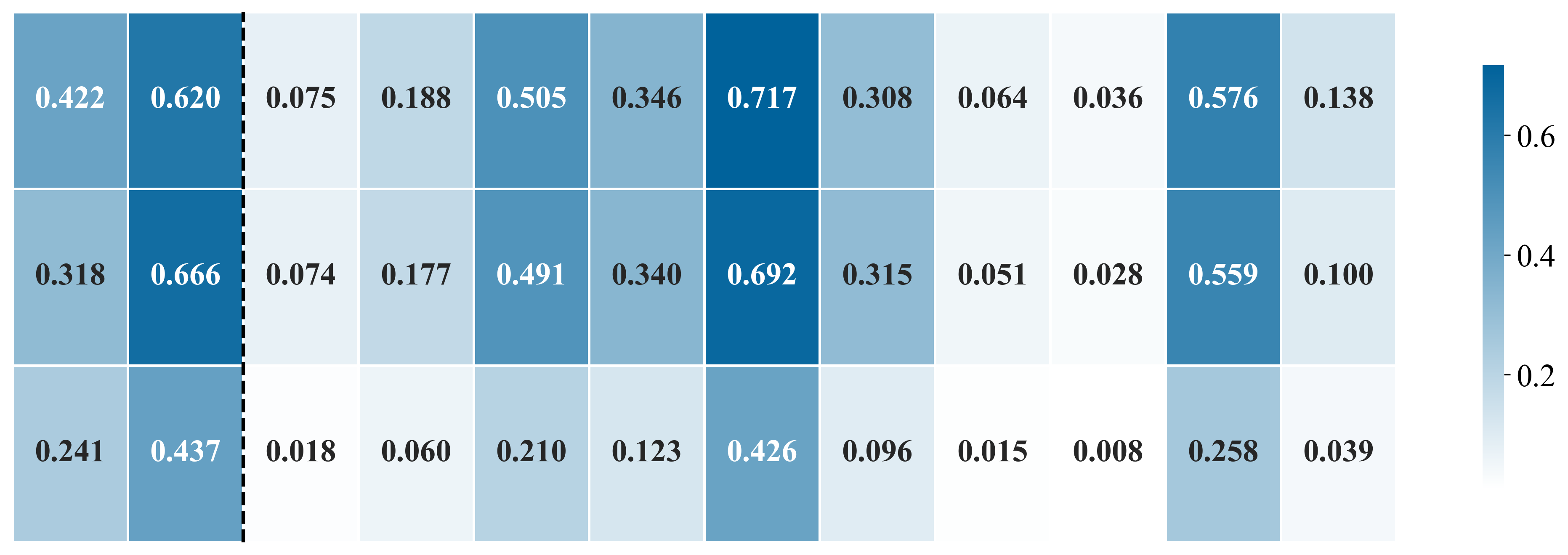}
        
        % --- X-Axis labels (12 columns: centers ~ 2, 9.5, 17, 24.5, 32, 39.5, 47, 54.5, 62, 69.5, 77, 84.5) ---
        \put(-1, -7){\rotatebox{45}{\scalebox{0.8}{\tiny YOLOv5$_{60e}$}}}
        \put(7, -7){\rotatebox{45}{\scalebox{0.8}{\tiny RT-DETRv4}}}
        \put(18, -3){\footnotesize \ding{172}}
        \put(25.5, -3){\footnotesize \ding{173}}
        \put(32.5, -3){\footnotesize \ding{174}}
        \put(40, -3){\footnotesize \ding{175}}
        \put(47.3, -3){\footnotesize \ding{176}}
        \put(54.5, -3){\footnotesize \ding{177}}
        \put(62, -3){\footnotesize \ding{178}}
        \put(69.5, -3){\footnotesize \ding{179}}
        \put(77, -3){\footnotesize \ding{180}}
        \put(84.5, -3){\footnotesize \ding{181}}

        % --- 纵轴 (Y-Axis) 设置 ---
        \put(-3, 5){\footnotesize \ding{182}}
        \put(-3, 16){\footnotesize \ding{183}}
        \put(-3, 28){\footnotesize \ding{184}}

    \end{overpic}
    
    \vspace{1em} 
    
    \caption{LVLMs performance in natural subset, benchmarked against the YOLOv5$_{60e}$ and the RT-DETRv4~\cite{lv2024detrs} detectors. LVLMs:~\ding{172} Grok 2;~\ding{173} Grok 4;~\ding{174} Gemini 2.5 Pro;~\ding{175} Gemini 2.5 Flash;~\ding{176} Gemini 3;~\ding{177} GPT 5;~\ding{178} Qwen;~\ding{179} Claude;~\ding{180} Doubao;~\ding{181} Hunyuan. Metrics:~\ding{182} mmAP$_{50:95}$;~\ding{183} mAR$_{50}$;~\ding{184} mAP$_{50}$.}
    \label{fig:natural-subset-performance}
\end{figure}

\autoref{fig:natural-subset-performance} summarizes detection performance on the natural subset, which consists of real-world footage with complex and unstructured visual noise. The strongest LVLM outperforms both baselines in $mAP_{50}$ and $mAR_{50}$. In particular, Gemini 3 (\ding{176}) achieves the best $mAP_{50}$ and $mAR_{50}$, exceeding YOLOv5$_{60e}$ and RT-DETRv4; its $mmAP_{50:95}$ remains comparable to RT-DETRv4 and well above YOLOv5$_{60e}$. Doubao (\ding{180}) and Gemini 2.5 Pro (\ding{174}) also outperform YOLOv5$_{60e}$ but remain below RT-DETRv4, indicating that several LVLMs maintain competitive detection quality in authentic driving scenes.

\subsubsection{Performance in Handcraft Subset}

\begin{figure}[ht]
    \centering
    \begin{overpic}[width=.95\columnwidth]{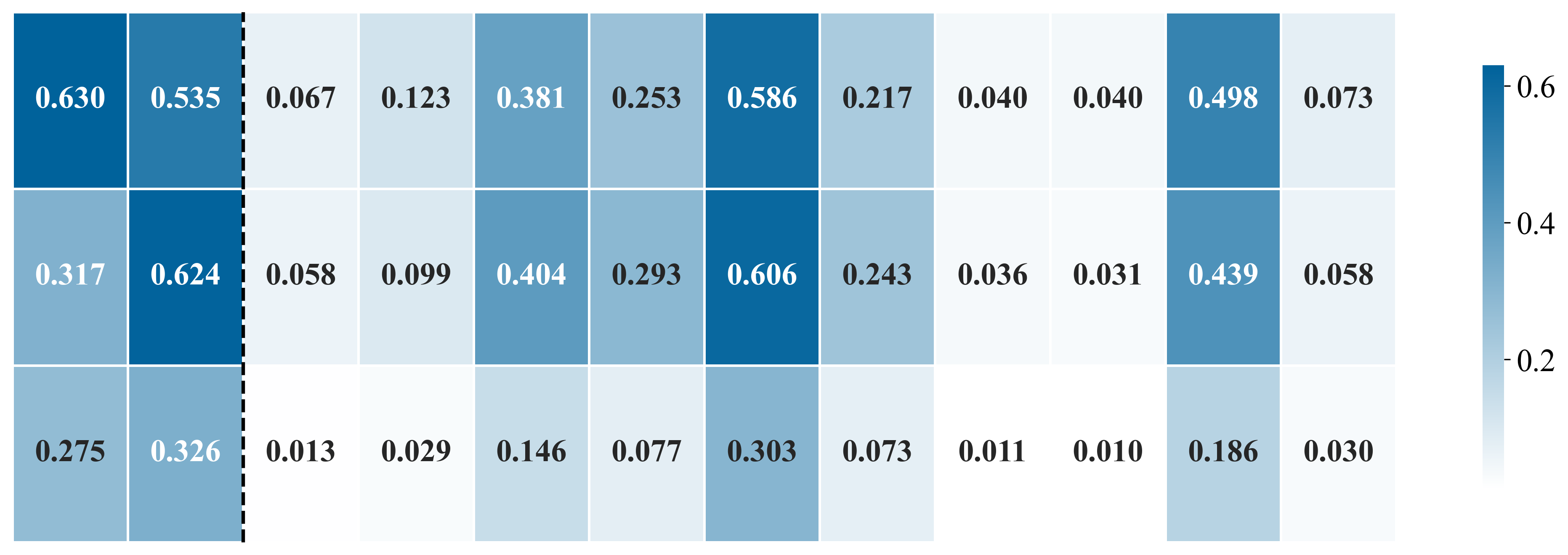}
        
        % --- X-Axis labels (12 columns: centers ~ 2, 9.5, 17, 24.5, 32, 39.5, 47, 54.5, 62, 69.5, 77, 84.5) ---
        \put(-1, -7){\rotatebox{45}{\scalebox{0.8}{\tiny YOLOv5$_{60e}$}}}
        \put(7, -7){\rotatebox{45}{\scalebox{0.8}{\tiny RT-DETRv4}}}
        \put(18, -3){\footnotesize \ding{172}}
        \put(25.5, -3){\footnotesize \ding{173}}
        \put(32.5, -3){\footnotesize \ding{174}}
        \put(40, -3){\footnotesize \ding{175}}
        \put(47.3, -3){\footnotesize \ding{176}}
        \put(54.5, -3){\footnotesize \ding{177}}
        \put(62, -3){\footnotesize \ding{178}}
        \put(69.5, -3){\footnotesize \ding{179}}
        \put(77, -3){\footnotesize \ding{180}}
        \put(84.5, -3){\footnotesize \ding{181}}

        % --- 纵轴 (Y-Axis) 设置 ---
        \put(-3, 5){\footnotesize \ding{182}}
        \put(-3, 16){\footnotesize \ding{183}}
        \put(-3, 28){\footnotesize \ding{184}}

    \end{overpic}
    
    \vspace{1em} 
    
    \caption{LVLMs performance in handcraft subset, benchmarked against the YOLOv5$_{60e}$ and the RT-DETRv4~\cite{lv2024detrs} detectors. LVLMs:~\ding{172} Grok 2;~\ding{173} Grok 4;~\ding{174} Gemini 2.5 Pro;~\ding{175} Gemini 2.5 Flash;~\ding{176} Gemini 3;~\ding{177} GPT 5;~\ding{178} Qwen;~\ding{179} Claude;~\ding{180} Doubao;~\ding{181} Hunyuan. Metrics:~\ding{182} mmAP$_{50:95}$;~\ding{183} mAR$_{50}$;~\ding{184} mAP$_{50}$.}
    \label{fig:handcraft-subset-performance}
\end{figure}

\autoref{fig:handcraft-subset-performance} reports the detection performance in the handcraft subset, which is composed of real-world images with manually constructed environment degradations. A different performance pattern is observed compared with the natural subset. The YOLOv5$_{60e}$ baseline achieves the highest $mAP_{50}$ in this subset, outperforming all LVLMs and the RT-DETRv4 detector.

In contrast, recall and $mmAP_{50:95}$ are dominated by RT-DETRv4, which surpasses YOLOv5$_{60e}$ and every LVLM on both metrics. Among the LVLMs, Gemini 3 (\ding{176}) comes closest to the transformer-based detector in recall, while Doubao (\ding{180}) still exceeds YOLOv5$_{60e}$ in recall, suggesting fewer missed detections than the anchor-based baseline. Overall, these results point to a trade-off: YOLOv5$_{60e}$ retains the strongest $mAP_{50}$ localization, whereas RT-DETRv4 and several LVLMs provide better object coverage on the handcraft subset.

\subsection{Performance in Object Subset}

\begin{figure}[ht]
    \centering
    \begin{overpic}[width=.95\columnwidth]{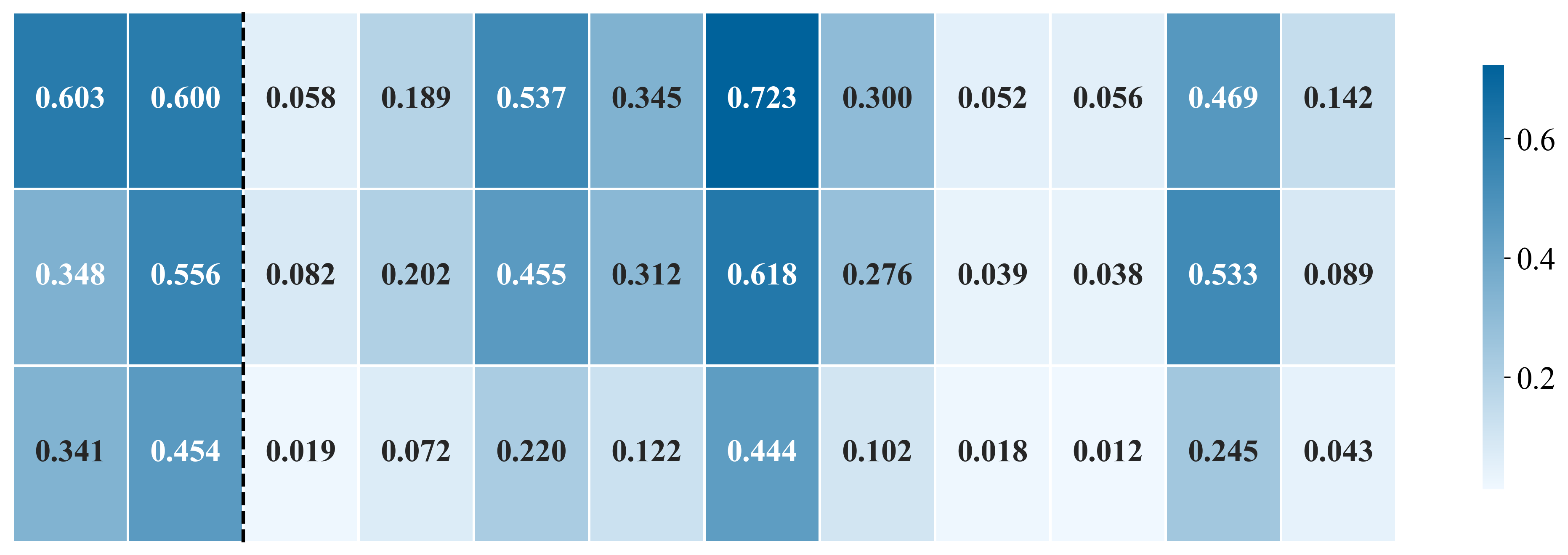}
        
        % --- X-Axis labels (12 columns: centers ~ 2, 9.5, 17, 24.5, 32, 39.5, 47, 54.5, 62, 69.5, 77, 84.5) ---
        \put(-1, -7){\rotatebox{45}{\scalebox{0.8}{\tiny YOLOv5$_{60e}$}}}
        \put(7, -7){\rotatebox{45}{\scalebox{0.8}{\tiny RT-DETRv4}}}
        \put(18, -3){\footnotesize \ding{172}}
        \put(25.5, -3){\footnotesize \ding{173}}
        \put(32.5, -3){\footnotesize \ding{174}}
        \put(40, -3){\footnotesize \ding{175}}
        \put(47.3, -3){\footnotesize \ding{176}}
        \put(54.5, -3){\footnotesize \ding{177}}
        \put(62, -3){\footnotesize \ding{178}}
        \put(69.5, -3){\footnotesize \ding{179}}
        \put(77, -3){\footnotesize \ding{180}}
        \put(84.5, -3){\footnotesize \ding{181}}

        % --- 纵轴 (Y-Axis) 设置 ---
        \put(-3, 5){\footnotesize \ding{182}}
        \put(-3, 16){\footnotesize \ding{183}}
        \put(-3, 28){\footnotesize \ding{184}}

    \end{overpic}
    
    \vspace{1em} 
    
    \caption{LVLMs performance in object subset of PeSOTIF, benchmarked against the YOLOv5$_{60e}$ and the RT-DETRv4~\cite{lv2024detrs} detectors. LVLMs:~\ding{172} Grok 2;~\ding{173} Grok 4;~\ding{174} Gemini 2.5 Pro;~\ding{175} Gemini 2.5 Flash;~\ding{176} Gemini 3;~\ding{177} GPT 5;~\ding{178} Qwen;~\ding{179} Claude;~\ding{180} Doubao;~\ding{181} Hunyuan. Metrics:~\ding{182} mmAP$_{50:95}$;~\ding{183} mAR$_{50}$;~\ding{184} mAP$_{50}$.}
    \label{fig:object-performance-subset}
\end{figure}

\autoref{fig:object-performance-subset} illustrates the detection performance heatmap on the object subset. Gemini 3 (\ding{176}) achieves the highest $mAP_{50}$ and $mAR_{50}$ across all evaluated models, while RT-DETRv4 attains a marginally higher $mmAP_{50:95}$, indicating tighter bounding-box localization across stricter IoU thresholds. The YOLOv5$_{60e}$ baseline exhibits competitive precision in terms of $mAP_{50}$ but substantially lower recall, highlighting its limited coverage under object-centric challenges. Gemini 3 (\ding{176}) demonstrates the most balanced capability across localization accuracy and detection coverage in the object subset.

\subsection{Ablation on the Ruler Overlay}
\label{sec-no-ruler}

\begin{figure}[ht]
    \centering
    \begin{overpic}[width=.95\columnwidth]{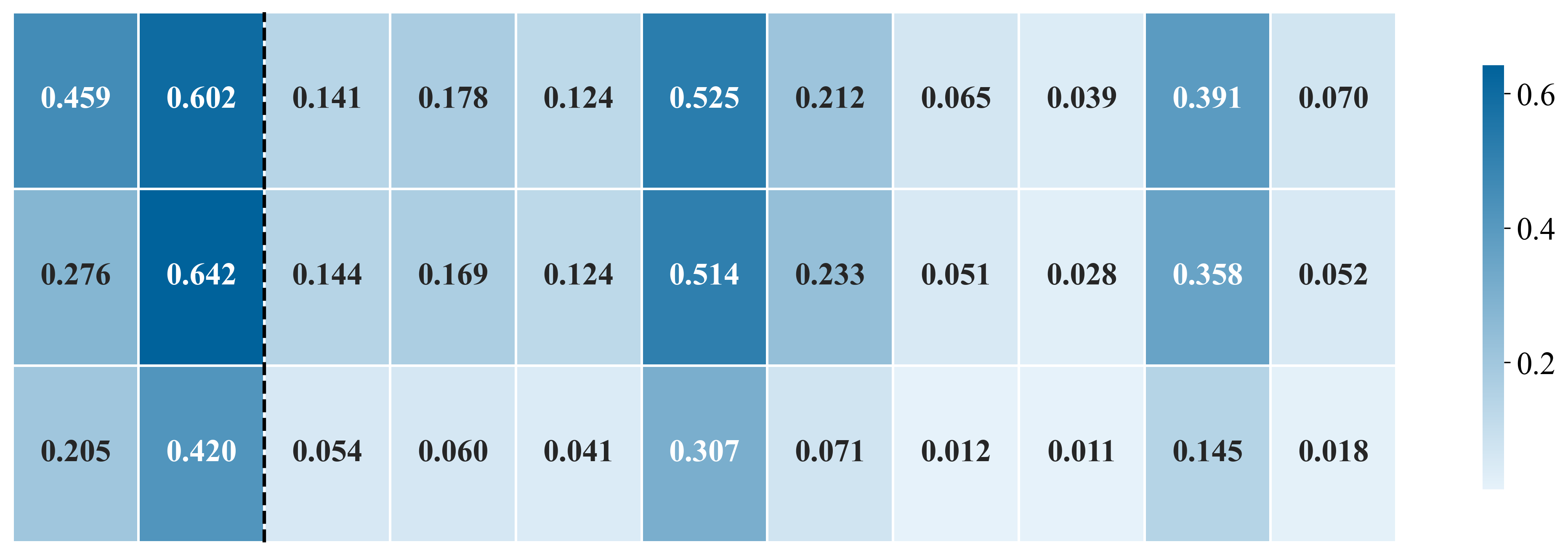}

        % --- X-Axis labels (11 columns: centers ~ 2, 10, 18, 27, 35, 43, 52, 60, 68, 77, 85) ---
        \put(-1, -7){\rotatebox{45}{\scalebox{0.8}{\tiny YOLOv5$_{60e}$}}}
        \put(7, -7){\rotatebox{45}{\scalebox{0.8}{\tiny RT-DETRv4}}}
        \put(17, -3){\footnotesize \ding{173}}
        \put(25.5, -3){\footnotesize \ding{174}}
        \put(34, -3){\footnotesize \ding{175}}
        \put(42.5, -3){\footnotesize \ding{176}}
        \put(51, -3){\footnotesize \ding{177}}
        \put(59.5, -3){\footnotesize \ding{178}}
        \put(68, -3){\footnotesize \ding{179}}
        \put(76.5, -3){\footnotesize \ding{180}}
        \put(85, -3){\footnotesize \ding{181}}

        % --- Y-Axis labels ---
        \put(-3, 5){\footnotesize \ding{182}}
        \put(-3, 16){\footnotesize \ding{183}}
        \put(-3, 28){\footnotesize \ding{184}}

    \end{overpic}

    \vspace{1em}

    \caption{Object-detection performance of LVLMs in PeSOTIF dataset when the ruler ticks are removed from the input. LVLMs:~\ding{173} Grok 4;~\ding{174} Gemini 2.5 Pro;~\ding{175} Gemini 2.5 Flash;~\ding{176} Gemini 3;~\ding{177} GPT 5;~\ding{178} Qwen;~\ding{179} Claude;~\ding{180} Doubao;~\ding{181} Hunyuan. Metrics:~\ding{182} mmAP$_{50:95}$;~\ding{183} mAR$_{50}$;~\ding{184} mAP$_{50}$.}
    \label{fig:no-ruler-performance}
\end{figure}

To substantiate the contribution of the ruler ticks introduced in the image preprocessing, an ablation study is conducted in which the ruler ticks are removed from the LVLM inputs while all other prompt components and image preprocessing operations remain unchanged. Grok 2 (\ding{172}) is excluded from this ablation as the model was discontinued and unavailable at the time of the ablation experiment. Two baselines, YOLOv5$_{60e}$ and RT-DETRv4, are unaffected by this ablation and are retained as references. The outcome is presented in~\autoref{fig:no-ruler-performance}.

A significant degradation is observed across all evaluated LVLMs. Gemini 3 (\ding{176}) remains the strongest LVLM but its $mAP_{50}$ drops from 0.695 to 0.525, and its $mmAP_{50:95}$ decreases from 0.407 to 0.307, indicating a clear loss in both detection coverage and localization tightness. Doubao (\ding{180}) exhibits a similar trend, with $mAP_{50}$ falling from 0.535 to 0.391 and $mmAP_{50:95}$ from 0.243 to 0.145. The remaining LVLMs, including Gemini 2.5 Pro (\ding{174}) and Gemini 2.5 Flash (\ding{175}), experience even larger relative drops, with all of them below 0.25 in $mAP_{50}$. In contrast, both baselines retain their original performance, and RT-DETRv4~\cite{lv2024detrs} outperforms all LVLMs across all metrics.

These observations confirm that the ruler ticks are essential for accurate object detection in the proposed pipeline. Without them, LVLMs must rely on implicit visual grounding alone, which yields noticeably weaker detection performance. This establishes the ruler overlay as a lightweight, training-free mechanism for bridging the gap between semantic reasoning and precise localization under SOTIF conditions.

\section{Conclusion and Discussion}
An evaluation of ten LVLMs for 2D object detection within SOTIF-relevant traffic scenarios is presented in this study. Utilizing the PeSOTIF dataset and a visual-prompting methodology, we demonstrated that general-purpose LVLMs can effectively translate high-level semantic understanding into structured spatial output. Very importantly, without any task-specific fine-tuning on PeSOTIF dataset, the leading LVLMs already surpass the YOLOv5$_{60e}$ baseline across all metrics, and exceed RT-DETRv4 in $mAP_{50}$ and $mAR_{50}$, while remaining slightly below in $mmAP_{50:95}$. The advantage is most pronounced in the natural subsets, where LVLMs exceed YOLOv5$_{60e}$ in recall by over 25\%, while closely matching RT-DETRv4 in the same subset, suggesting that their global contextual reasoning is more resilient to the degradation of local visual features than the feature-matching pipelines of YOLOv5$_{60e}$ and RT-DETRv4.

The ablation study in Section III-\ref{sec-no-ruler} demonstrates the value of the ruler ticks as a core component of the proposed methodology. When the ruler is removed, the geometric accuracy of every evaluated LVLM degrades and RT-DETRv4~\cite{lv2024detrs} regains the overall lead; when the ruler is provided, leading LVLMs move ahead of both baselines. This contrast establishes the ruler overlay as a lightweight, training-free prompt-level design that unlocks the geometric localization capability of pretrained LVLMs without any model modification. The ruler is therefore not an incidental trick but a principled and transferable interface between general-purpose LVLMs and precise geometric tasks.

% This contrast establishes the ruler overlay as a lightweight yet decisive prompt-level design: with a single, training-free modification of the input, the latent spatial reasoning of pretrained LVLMs is converted into usable geometric output, enabling them to outperform a fully trained state-of-the-art transformer detector without any modification to the model itself. The ruler is therefore not an incidental trick but a principled and transferable interface between general-purpose LVLMs and dense geometric tasks.

Despite the promising detection capabilities, practical deployment faces challenges regarding computational efficiency, as the current processing speed of LVLMs generally lags behind the real-time requirements of automated driving. Consequently, LVLMs are currently best positioned as a high-level \enquote{safety validator} or a redundant perception branch to handle long-tail corner cases that challenge conventional sensors. Future research should target the distillation of these LVLMs into lightweight architectures and the refinement of end-to-end spatial alignment to further bridge the gap between semantic reasoning and real-time execution.

% \section*{Conflict of Interest}
% One of the authors, Prof. Johannes Betz, currently serves as an Associate Editor for IEEE Open Journal of Intelligent Transportation Systems. To avoid any conflict of interest, Prof. Johannes Betz will not be involved in the review or editorial decision process of this manuscript.

\bibliographystyle{IEEEtran}
\bibliography{ref}

\end{document}